# Evidential Reasoning in Parallel Hierarchical Vision Programs

Ze-nian Li[**] and Leonard Uhr

*Department of Computer Sciences*
*University of Wisconsin-Madison*

ABSTRACT

This paper presents an efficient adaptation and application of the Dempster-Shafer theory of evidence, one that can be used effectively in a massively parallel hierarchical program for visual pattern perception. It describes the techniques used, and shows in an extended example how they serve to improve the program's performance as it applies a multiple-level set of processes.

## 1. Introduction

We have been developing programs for visual perception which can be executed in a pyramid-like structure that consists of many thousands of computers. The hierarchy of micro-modular transforms executed by this parallel network of processors has been effective for analyzing images containing complex objects like houses and neurons [Li86a, Li86b, Li85, Uhr79]. This paper describes a new evidential reasoning mechanism based on the Dempster-Shafer theory of evidence that we have recently developed for and incorporated into these pyramid vision programs.

The vision program uses *key features* and *evidential reasoning*. Initially, low level local features are assessed in parallel by low level processors, and then used to compute more global and abstract features at consecutively higher levels in the pyramid. Once certain important 'key' features are extracted, a hypothesis will be generated. Typically the program will then move, in a top-down manner, to lower levels in order to search for more evidence that might verify or deny the hypothesis. The reasoning processes can thus be executed in a parallel and hierarchical manner within different nodes and different levels in the pyramid.

Basically, the evidential reasoning is attempting two tasks: *evidence accumulation* and *hypothesis verification*. *Uncertainty* and *incompleteness* are inevitable in both phases. As is the case with many computer vision systems, our programs are compelled to make decisions on uncertain and incomplete evidence and knowledge. This is especially true whenever factors such as the achievable quality of the image data, the complexity of the domain, and the computational cost, are taken into account.

The new evidential reasoning mechanism is designed to achieve the following goals:
(a) Accommodating uncertainty and incompleteness, both in the evidence and in the world knowledge.
(b) Making the knowledge representation scheme flexible, and easy to update.
(c) Emphasizing the role of some key features and, when desirable, conjunctions of features that are important for the object's recognition.
(d) Embedding the reasoning mechanisms into the basic format of the micro-modular transforms, so they can be executed in parallel in the pyramid.

Section 2 presents a set-theoretical evidential reasoning approach. The method for evidence accumulation is illustrated. A new knowledge representation technique is introduced, and hence a new mechanism for hypothesis verification. In section 3, a program for window recognition will be described. The results obtained indicate that the evidential reasoning mechanism yields good results. Section 4 concludes the paper.

---

[*] This research was supported by NSF Grant DCR-8302397 to L. Uhr.
[**] Ze-nian Li is with the Department of Electrical Engineering and Computer Science, University of Wisconsin-Milwaukee.

175

## 2. A Set-Theoretical Evidential Reasoning Approach to Computer Vision

In this paper, a set-theoretical approach based on the Dempster-Shafer mathematical theory of evidence will be introduced. The description of the Dempster-Shafer theory can be found in [Demp67, Shaf76]. Here, the notation in [Shaf76] is adopted.

Let $\Theta$ be a set of propositions about the mutually exclusive and exhaustive possibilities in a domain. $\Theta$ is called the *frame of discernment* and $2^\Theta$ is the set of all possible subsets of $\Theta$.

A function $m: 2^\Theta \rightarrow [0, 1]$ is called a *basic probability assignment* if it satisfies $m(\varnothing) = 0$ and $\sum_{A \subseteq \Theta} m(A) = 1$.

If $m$ is a basic probability assignment, then a *belief function* is defined by
$$Bel(A) = \sum_{B \subseteq A} m(B), \quad \text{for } A \subseteq \Theta.$$

### 2.1. Evidence Accumulation

The accumulation of evidence is conducted in two steps, for single feature assessment and for multi-feature combination.

**Single Feature Assessment**

At this step, probability values are assigned to the extracted features to represent their uncertainty and incompleteness. The following factors are taken into account:

(1) The quality of the input data, and the weights of the extracted features.

(2) The 'goodness' of feature values. A feature value is compared to the typical feature value of a hypothesized object to answer questions like "How good is the shape of this *long* region, if it is compared to the shape of a typical window shutter". As suggested by Ragade and Gupta [Raga77], many functions $0 \leq \mu(v, M) \leq 1$ can be used to evaluate the feature values, where $v$ denotes the extracted feature value, $M$ is the typical feature value of a hypothesized object and $M_1$, $M_2$ and $t$ are constants. The following is the step function that is used in our program. Apparently, many alternatives and variations of these functions can be invented.

Step Function:  
$\mu(v) = 1 \qquad M - M_1 \leq v \leq M + M_1$  
$\mu(v) = t \quad (0 < t < 1) \qquad M - M_2 \leq v < M - M_1 \text{ or } M + M_1 < v \leq M + M_2$  
$\mu(v) = 0 \qquad \text{otherwise.}$

The factors in (1) and (2) are combined into a single probability value. As a result, each feature $f$ gets a probability mass $0 \leq m(f) \leq 1$. In most cases, $m(f)$ is less than 1, the remainder of the unit mass $1 - m(f)$ is assigned to $\Theta$ to represent the uncertainty. Therefore, for single feature assessment simple belief functions are used, where $Bel(f) = m(f)$, and $Bel(\Theta) = m(\Theta)$.

**Multi-feature Combination**

If each extracted feature is viewed as a piece of evidence, then the independent observations on multi-features can be combined to serve as accumulated evidence. Belief functions of independently extracted features are combined using Dempster's combination rule.

Suppose $m_1$ is the basic probability assignment for a belief function $Bel_1$ over a frame $\Theta$, with focal elements $A_i$ ($i = 1, \cdots, k$); and similarly $m_2$ is the basic probability assignment for a belief function $Bel_2$ over the same frame $\Theta$, with focal elements $B_j$ ($j = 1, \cdots, l$).

If $K = \sum_{A_i \cap B_j = \varnothing} m_1(A_i) m_2(B_j) < 1$, then the function $m: 2^\Theta \rightarrow [0, 1]$ is a basic probability assignment, defined by:

(1) $m(\varnothing) = 0$, and

(2) $m(C) = (1 - K)^{-1} \sum_{A_i \cap B_j = C} m_1(A_i) m_2(B_j)$.

The belief function, $m$, that combines $Bel_1$ and $Bel_2$, is called the *orthogonal sum* of $Bel_1$ and $Bel_2$, denoted

176

$Bel_1 \oplus Bel_2$. Since Dempster's combination rule is associative and commutative, features can be combined in any order.

## 2.2. A New Knowledge Representation Technique

In the past few years, the Dempster-Shafer theory has been proposed in several application domains. Some of the work includes [Garv81, Lowr82, Wesl84]. In AI systems, a sound knowledge representation scheme is crucial to the success of such reasoning systems.

In the field of computer vision, one of the popular knowledge representation schemes used in the evidential reasoning is the *Dependency-graph model* [Lowr82]. The assumption is that the domains of interest can be modeled in a propositional framework. The frame of discernment is represented by dependency-graphs, in which the nodes represent propositions and the arcs represent relationships, such as $and$, $or$, $\rightarrow$, $not$, etc., between the propositions. Based on Dempster-Shafer's evidential reasoning theory, Lowrance claimed that a formal system capable of pooling and extending evidential information could be built, while maintaining internal consistency.

The dependency-graph model approach has been successfully demonstrated in several domains that contain small sets of objects. It has the merit of being formal and sound. However it has drawbacks. The success of the dependency-graph model relies on the construction of a fairly complete graph. Due to the complexity of the graph, the world model would take potentially very large amounts of space to store, and time to process, and it would be difficult to update. As stated in [Lowr82], "as these graphs become larger and more complex, it becomes increasingly difficult to guarantee their logical consistency."

The essential point is that a knowledge representation scheme should be able to utilize incomplete knowledge. There should be a way to emphasize the important subset of the world knowledge. Truth should be expressed relative to portions of the environment, sensitive to the context. Although the initial model will probably generate poor results, it should be flexible and easy to improve.

Our new knowledge representation technique is characterized by associating probability values to the expected feature components of the hypothesized objects. Probability values are used to answer questions like "If it has a typical elongated shape for a window shutter, how much should this feature of shape contribute to the belief for a shutter ?"

To discriminate different objects in a vision system, it is often important to consider beliefs about multiple features, especially those key features of the hypothesized object. If the knowledge about window shutters is that "Window shutters are usually elongated rectangular regions, with a low 'edgeness' measure, next to windows", then the following probability mass assignments $m_s$ (*shutter*) may be given to the window shutter:

$$m\ (long) = 0.25,\ m\ (low) = 0.15,\ m\ (long\ \&\ low) = 0.15,\ m\ (next-to) = 0.25,\ m\ (\Theta) = 0.2.$$

The importance of the feature is represented by the amount of the mass assigned to it. The numbers initially assigned are more or less subjective and 'arbitrary'. But they can be adjusted later on, either by the programmer or by a learning program.

If the current task is to discriminate shutters from a chimney, and the system's knowledge about chimneys is that "Chimneys are usually elongated rectangular regions, not next-to windows, and no knowledge about the importance of the texture", then a mass assignment $m_s$ (*chimney*) may be given at the same time,

$$m\ (long) = 0.25,\ m\ (next-to) = 0.35,\ m\ (\Theta) = 0.4.$$

In this way, the knowledge representation technique allows the invocation of the subsets of the world knowledge that are most closely related to the current context of the recognition, and have the most discrimination power for the current hypotheses.

## 2.3. Hypothesis Verification

The mechanism of hypothesis verification can be viewed as a set of mappings between two spaces. The first is the evidence space $E$, and the second the object space $O$. A mapping

$$e \rightarrow o$$

defines a belief function $Bel\ (o)$ over the object space $O$, where $e \subset E$ is the set of accumulated evidence, $o \subset O$ is the hypothetical object, and $0 \leq Bel\ (o) \leq 1$ is the belief committed to $o$. While conducting this mapping, the knowledge sources are consulted.

As described in section 2.1 and 2.2, the accumulated evidence can be represented by a belief function whose mass distribution is $m_e$, whereas the system's knowledge of the hypothetical object can be represented by another belief function whose mass distribution is $m_s$. We here introduced a third belief function $Bel(o)$ to represent the result of the hypothesis verification.

(a) $Bel(o) = \sum_{\substack{A_i \subseteq B_j \\ B_j \neq \Theta}} m_e(A_i) \, m_s(B_j),$

(b) $Bel(\Theta) = 1 - Bel(o).$

This belief function is graphically depicted in Fig. 1. The areas whose masses are committed to the hypothetical object $o$ are marked with *'s. The value of $Bel(o)$ is the sum of these areas. The belief function thus generated is a simple belief function. When the evidential reasoning is applied hierarchically, $Bel(o)$ can be used together with other pieces of evidence for the higher level object it implies.

Fig. 2 illustrates this type of hypothesis verification. The knowledge of window shutters ($m_s(shutter)$) described in section 2.2 is used. Some accumulated evidence $m_e$ is assumed. The new generated belief function of this hypothesis verification is

$Bel(shutter) = 0.21 \times (0.15 + 0.15 + 0.25 + 0.25) + 0.14 \times (0.15 + 0.25) + 0.09 \times (0.25 + 0.25) +$
$\qquad 0.06 \times 0.25 + 0.21 \times (0.15 + 0.15 + 0.25) + 0.14 \times 0.15 + 0.09 \times 0.25 = 0.44.$
$Bel(\Theta) = 1 - 0.44 = 0.56.$

### 3. Programming Results

Experimental evaluations have given satisfactory results with four images of houses and office buildings. The window shutter example in the previous section actually comes from Fig. 3(a). In this section, Fig. 3(b), the image of an office building, will be used as an example to illustrate how a program that is capable of recognizing windows of buildings uses our evidential reasoning mechanism. The original digitized image has a resolution of 512 x 512 pixels. For the examples demonstrated in this section, a reduced resolution 128 x 128 is used; it is stored at level 7 of the pyramid.

The algorithms for recognizing houses can be found in [Li85]. Briefly, the program first locates micro-edges with eight possible directions at level 7. Next, short edges are extracted by level 6 nodes, and long edges by level 5 nodes. Pairs of parallel long edges with opposite directions (e.g., long edges with 0° and 180°, or 90° and 270°) have been found to be good clues for windows with rectangular shapes. Thus pairs of such horizontal long edges are used to locate the possible window areas. Fig. 4 shows all the possible window areas located in the office building image. (Forty six possible window areas are hypothesized by the program. For ease of demonstration, the real window areas are labeled W1 – W12, other (false) areas are labeled by number, 1 – 34. Since the procedure being examined looks for only horizontal long edges, six narrow windows in Fig.

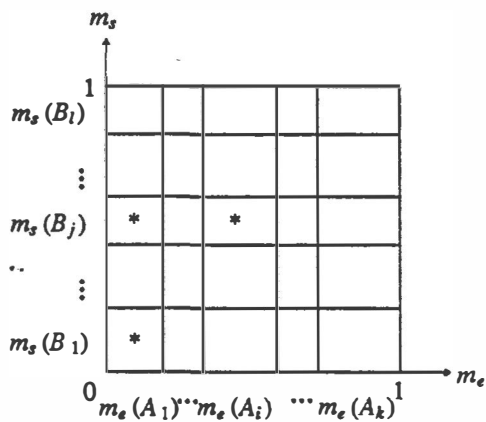

Fig. 1 Belief Function $Bel(o)$

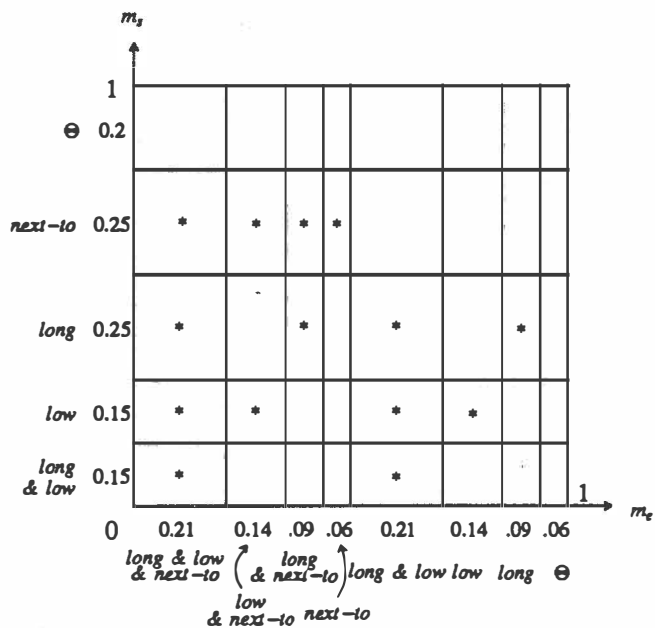

Fig. 2 Hypothesis Verification

3(b) are not located in this step.)

Apparently additional evidence is needed and should be accumulated, to better distinguish the true window areas from others. The following are illustrations of how multiple features (evidence) can be employed to help the process of evidential reasoning.

**Evidence of Shapes and Texture**

Several features of the possible window areas are assessed at level 5.

(1) The elongation of the area. The *elongation* of the rectangular areas is defined as *length / width*, if *length* $\geq$ *width*; *width / length*, otherwise. Under this definition, the elongation values of typical windows are likely to be small positive numbers larger than one. Thus the following belief values are given for the measured elongation.

$$Bel\ (elong) = \begin{cases} 0.5, & \text{if } elongation \leq 3; \\ 0.3, & \text{if } 3 < elongation \leq 5; \\ 0, & \text{otherwise.} \end{cases}$$

(2) The texture of the area. The interior texture of window areas can often be measured by the number of micro-edges. For opened windows or big glass windows, there are very few micro-edges inside the areas. In case of windows with a few panes or half-closed window shades, there will be overwhelmingly more horizontal and vertical edges (micro-edge 0, 2, 4, 6) than diagonal edges (micro-edge 1, 3, 5, 7). Hence the belief value for the texture measure is

$$Bel\ (text) = \begin{cases} 0.4, & \text{if } edgedness < 0.1; \\ 0.4, & \text{if } hv/d \geq 4; \\ 0.2, & \text{if } 2 \leq hv/d < 4; \\ 0, & \text{otherwise.} \end{cases}$$

where *edgedness* = number-of-edges / area-of-region, and $hv/d$ is the ratio of the total number of the horizontal and vertical edges to the number of diagonal edges. The micro-edges are extracted at level 7. The statistical measures are implemented by pyramid operations.

(3) The vertical edges that serve as the left and right boundaries of the windows. Perfect boundaries cannot always be extracted. $Bel$ $(left-bound)$ and $Bel$ $(right-bound)$ are thus used in a similar way to assess the evidence of the existence of boundaries.

The belief functions initially assigned to the single features are all 'simple belief functions'. While $Bel$ is assigned to the proposition (e.g., $elong$) supporting the window hypothesis, the value of $1 - Bel$ is assigned to $\Theta$ to represent the uncertainty.

After all four features have been assessed individually, these simple belief functions are combined to derive a new belief function:
$$Bel\ (elong) \oplus Bel\ (text) \oplus Bel\ (left-bound) \oplus Bel\ (right-bound),$$
whose mass distribution is denoted by $m_e$ in section 2.2. For verifying window hypotheses, the combined evidence is compared to the knowledge source, having the following probability mass distribution, $m_{s_1}$:
$$m\ (elong) = 0.15,\ m\ (text) = 0.20,\ m\ (bound) = 0.35,\ m\ (\Theta) = 0.3.$$
The existence of boundaries is thought to be the most convincing evidence, thus a big portion of the total mass is assigned to the proposition $bound$. A certain amount of the mass is attributed to $\Theta$ to reflect the incompleteness of this knowledge source.

The results of this verification are the belief functions $Bel\ (window)$ for the hypothesized window areas. The belief values for single features and $Bel\ (window)$ are shown in Table 1. The 18 possible window areas in Table 1 are the ones that obtain higher $Bel\ (window)$ values. The rest 28 (less possible) areas are not shown.

### Evidence about the Geometrical Relations of Windows

The operations just described succeed in yielding higher belief values for the 12 real windows than for any of the false windows. At this step, the geometrical relations between windows are used to further enhance the results. Usually the windows of a building are arranged in a horizontal or vertical alignment. To carry out the necessary 'lateral search' for appropriately positioned neighboring windows, higher level nodes in the pyramid are used. Belief values $Bel\ (window)$ at level 5 are passed one level up to the nodes at level 4. All the surviving nodes will search for possible sibling window areas vertically and horizontally. In case a vertical sibling is found, the $Bel\ (v-sibl)$ will be set to 0.6. In the same way, $Bel\ (h-sibl)$ will be set. Consequently, the combined belief function $Bel\ (window) \oplus Bel\ (v-sibl) \oplus Bel\ (h-sibl)$ is checked with the knowledge source for verifying the hypothesis that the window area in question is one of the windows in a building. The knowledge source is represented by $m_{s_2}$:
$$m\ (window) = 0.4,\ m\ (v-sibl) = 0.2,\ m\ (h-sibl) = 0.2,\ m\ (v-sibl\ \&\ h-sibl) = 0.2.$$
A part of the mass values is assigned to the conjunction of $v-sibl$ and $h-sibl$ to emphasize the chance of the coexistence of both siblings. Because this piece of knowledge is judged to be very certain, no mass is assigned to $\Theta$.

The resulting belief values are $Bel'\ (window)$ in Table 1. The previous results have been enhanced by the usage of the geometrical relations between expected windows. While the belief values for non-window areas decrease (from 0.335 to 0.134 for area 4 and 15), the values for all the real window areas get increased. Thus the contrast between the window and non-window areas becomes better.

### Example of Conflicting Evidence

One of the advantages of the Dempster-Shafer set-theoretical theory of evidence is the ability to accommodate conflicting evidence. The last part of our example will illustrate this capability, by applying the reasoning mechanism to independent observations of conflicting evidence.

The simplicity of the image of a building being used, e.g., the flat roof and the clear outline, makes it possible for the program to find the building's boundaries. Any areas falling outside the boundaries could hardly be windows of the building. Therefore a proposition $non-window$ is introduced as the negation of the proposition $window$. Areas outside of the building boundaries are assigned $Bel\ (non-window) = 0.5$, other areas will be assigned $Bel\ (non-window) = 0$. The remainder of the unit mass is assigned to $\Theta$. The new belief about $non-window$ is combined with the belief about $window$, $v-sibl$ and $h-sibl$ to derive
$$Bel\ (non-window) \oplus Bel\ (window) \oplus Bel\ (v-sibl) \oplus Bel\ (h-sibl).$$
Finally, the same knowledge source with $m_{s_2}$ is utilized to verify the $windows$ hypothesis. As expected, the belief values for those 'outside' nodes are further decreased (See $Bel''\ (window)$ in Table 1).

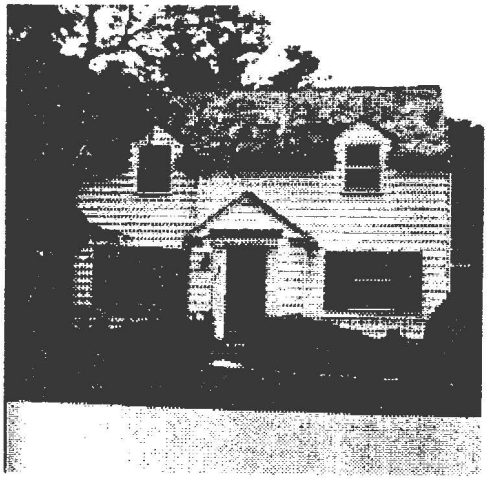
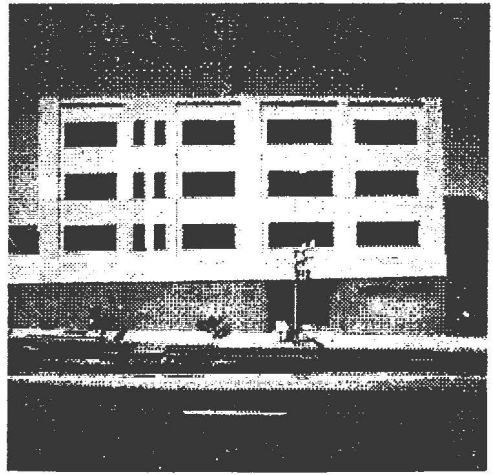

(a) A House Image    (b) An Office Building Image

Fig. 3 Images of a House and an Office Building

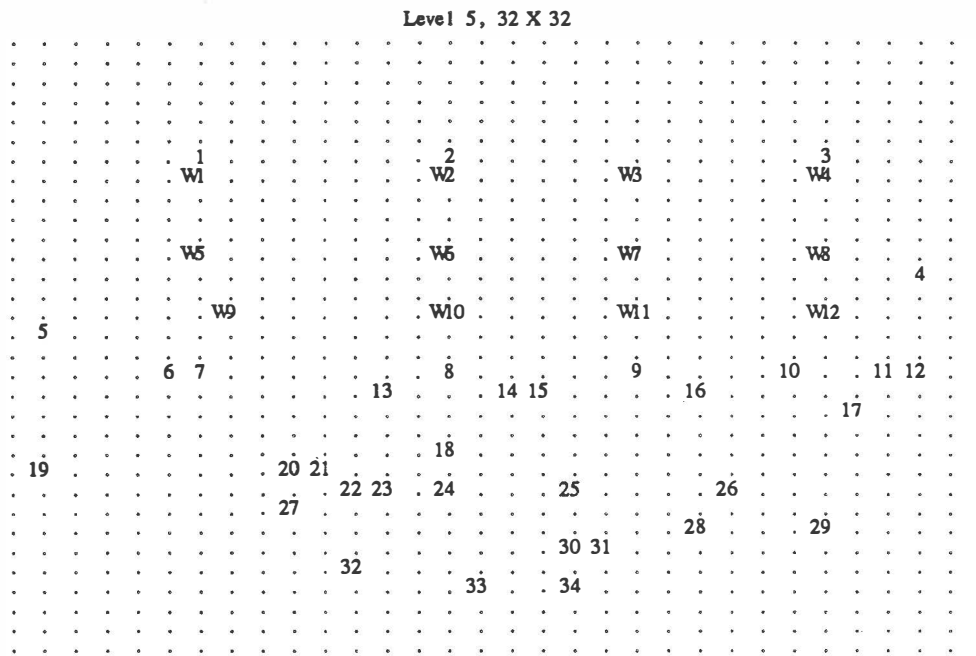

Fig. 4 Possible Window Areas

|  | Possible window areas | | | | | | | | | | | |
|---|---|---|---|---|---|---|---|---|---|---|---|---|
|  | W1-6 | W7 | W8 | W9 | W10 | W11 | W12 | 4 | 5 | 9 | 15 | 17 | 18 |
| *Bel (elong)* | 0.5 | 0.5 | 0.5 | 0.5 | 0.5 | 0.5 | 0.5 | 0.3 | 0.5 | 0.5 | 0.3 | 0.5 | 0.5 |
| *Bel (text)* | 0.4 | 0.2 | 0.4 | 0.4 | 0.4 | 0.4 | 0.4 | 0.4 | 0 | 0.4 | 0.4 | 0 | 0 |
| *Bel (lt−bound)* | 0.6 | 0.6 | 0.6 | 0.6 | 0.6 | 0.6 | 0.6 | 0 | 0.1 | 0 | 0 | 0.6 | 0.3 |
| *Bel (rt−bound)* | 0.6 | 0.6 | 0.6 | 0.3 | 0.1 | 0.6 | 0.6 | 0.6 | 0.3 | 0.3 | 0.6 | 0 | 0.1 |
| *Bel (wnd)* | .449 | .409 | .449 | .407 | .379 | .449 | .449 | .335 | .205 | .262 | .335 | .285 | .205 |
| *Bel (v−sibl)* | 0.6 | 0.6 | 0.6 | 0.6 | 0.6 | 0.6 | 0.6 | 0 | 0 | 0.6 | 0 | 0.6 | 0.6 |
| *Bel (h−sibl)* | 0.6 | 0.6 | 0.6 | 0.6 | 0.6 | 0.6 | 0.6 | 0 | 0.6 | 0 | 0 | 0 | 0 |
| *Bel ' (wnd)* | .492 | .475 | .492 | .475 | .462 | .492 | .492 | .134 | .203 | .225 | .134 | .234 | .203 |
| *Bel (non−wnd)* | 0 | 0 | 0 | 0 | 0 | 0 | 0 | 0.5 | 0.5 | 0 | 0 | 0 | 0.5 |
| *Bel " (wnd)* | .492 | .475 | .492 | .475 | .462 | .492 | .492 | .080 | .166 | .225 | .134 | .234 | .166 |

Table 1 Belief Values for Possible Windows

### 4. Conclusion

This paper describes how the Dempster-Shafer theory of evidence is used in a massively parallel hierarchical pyramid structure for computer vision. The program used was originally designed to recognize complex real-world objects (e.g., houses, office buildings, neurons) using a structure of micro-modular production rule-like transforms that are applied in a combined bottom-up top-down flow. Evidential reasoning was embedded at several stages in this program's processes. As shown by our examples, it serves to disambiguate and enhance the program's judgments about objects. Preliminary tests indicate that it improved the program's performance. The knowledge representation scheme described in this paper uses incomplete world knowledge. It is flexible and easy to update. Its reasoning mechanism extends the applications of the belief functions and Dempster's combination rule in a relatively efficient and powerful manner.